\documentclass[twoside,11pt]{article}

\usepackage{jmlr2e}

\usepackage[utf8]{inputenc} 
\usepackage[T1]{fontenc}    
\usepackage{hyperref}       
\usepackage{url}            
\usepackage{booktabs}       
\usepackage{amsfonts}       
\usepackage{nicefrac}       
\usepackage{microtype}      
\usepackage{comment}        
\usepackage{amsmath}
\usepackage{multirow}
\usepackage{bm}
\usepackage{hhline}
\usepackage{amssymb}
\usepackage{makecell}
\usepackage{mathtools}
\usepackage{color}
\usepackage{xcolor}
\usepackage{MnSymbol}
\usepackage{makecell}
\usepackage{graphicx}
\usepackage{arydshln}
\usepackage{booktabs}
\usepackage{framed}
\usepackage[font=small]{caption}
\usepackage[scientific-notation=true]{siunitx}
\usepackage{wrapfig}
\usepackage{algorithm}
\usepackage{algorithmic}
\usepackage{lipsum}
\usepackage{sidecap}
\usepackage{pifont}
\usepackage{fixltx2e}
\usepackage[flushleft]{threeparttable}
\usepackage{diagbox}
\usepackage{listings}

\newcommand{\cmark}{\ding{51}}%
\newcommand{\xmark}{\ding{55}}%

\newcommand{\names}{\textsc{MultiBench}}

\newcommand{\codes}{\textsc{MultiZoo}}

\definecolor{gg}{RGB}{15,150,15}
\definecolor{rr}{RGB}{230,45,45}

\makeatletter
\def\maketag@@@#1{\hbox{\m@th\normalfont\normalsize#1}}
\makeatother

\usepackage{amsmath,amsfonts,bm}

\def\eqref#1{eq~(\ref{#1})}

\def\1{\bm{1}}

\DeclareMathAlphabet{\mathsfit}{\encodingdefault}{\sfdefault}{m}{sl}
\SetMathAlphabet{\mathsfit}{bold}{\encodingdefault}{\sfdefault}{bx}{n}

\usepackage{hyperref}

\usepackage{booktabs}       %
\usepackage{amsfonts}       %
\usepackage{nicefrac}       %
\usepackage{microtype}      %
\usepackage{subcaption}

\usepackage{enumitem}
\setlist{nolistsep}
\setlist[itemize]{noitemsep, topsep=0pt}

\usepackage{times}

\usepackage{graphicx} %
\usepackage{color}

\usepackage{array}
\usepackage{amssymb}
\usepackage{amsmath}
\usepackage{xspace}
\usepackage{fancyhdr}
\usepackage{comment}

\newcolumntype{H}{>{\setbox0=\hbox\bgroup}c<{\egroup}@{}}

\newcommand{\noaistats}[1]{}  %

\definecolor{darkgreen}{rgb}{0,0.4,0.0}
\definecolor{darkblue}{rgb}{0,0.1,0.3}
\definecolor{darkred}{rgb}{0.7,0.0,0.0}

\AtBeginDocument{%
  \addtolength\abovedisplayskip{-0.3\baselineskip}%
  \addtolength\belowdisplayskip{-0.3\baselineskip}%
}

\usepackage{lastpage}
\jmlrheading{24}{2023}{1-\pageref{LastPage}}{9/22; Revised
4/23}{5/23}{22-1021}{Paul Pu Liang, Yiwei Lyu, Xiang Fan, Arav Agarwal, Yun Cheng, Louis-Philippe Morency, Ruslan Salakhutdinov}

\ShortHeadings{\codes\ \& \names: A Toolkit for Multimodal Learning}{Liang, Lyu, Fan, Agarwal, Cheng, Morency, Salakhutdinov}

\firstpageno{1}

\begin{document}

\title{\codes\ \& \names:\\A Standardized Toolkit for Multimodal Deep Learning}

\author{\name Paul Pu Liang, Yiwei Lyu, Xiang Fan, \email \{pliang,ylyu1,xiangfan\}@cs.cmu.edu \\
        \name Arav Agarwal, Yun Cheng, \email \{arava,yuncheng\}@cs.cmu.edu \\
        \name Louis-Philippe Morency, Ruslan Salakhutdinov \email \{morency,rsalakhu\}@cs.cmu.edu \\
        \addr Machine Learning Department and Language Technologies Institute, Carnegie Mellon University}%
\vspace{-6mm}
\editor{Antti Honkela}

\maketitle

\vspace{-10mm}
\begin{abstract}
Learning multimodal representations involves integrating information from multiple heterogeneous sources of data.
In order to accelerate progress towards understudied modalities and tasks while ensuring real-world robustness, we release \codes, a public toolkit consisting of standardized implementations of $>20$ core multimodal algorithms and \names, a large-scale benchmark spanning $15$ datasets, $10$ modalities, $20$ prediction tasks, and $6$ research areas. Together, these provide an automated end-to-end machine learning pipeline that simplifies and standardizes data loading, experimental setup, and model evaluation. To enable holistic evaluation, we offer a comprehensive methodology to assess (1) generalization, (2) time and space complexity, and (3) modality robustness.
\names\ paves the way towards a better understanding of the capabilities and limitations of multimodal models, while ensuring ease of use, accessibility, and reproducibility. Our toolkits are publicly available, will be regularly updated, and welcome inputs from the community\footnote{\names\ was previously published at NeurIPS 2021~\citep{liang2021multibench}, although the datasets and algorithms were the central contributions of that publication, not the software. This paper focuses on the open-source software along with a larger collection of datasets, algorithms, and evaluation metrics.}.

Code: \url{https://github.com/pliang279/MultiBench}

Documentation: \url{https://multibench.readthedocs.io/en/latest/}
\end{abstract}


\vspace{-3mm}
\section{Introduction}
\vspace{-2mm}

\begin{figure*}[tbp]
\centering
\vspace{-0mm}
\includegraphics[width=0.9\linewidth]{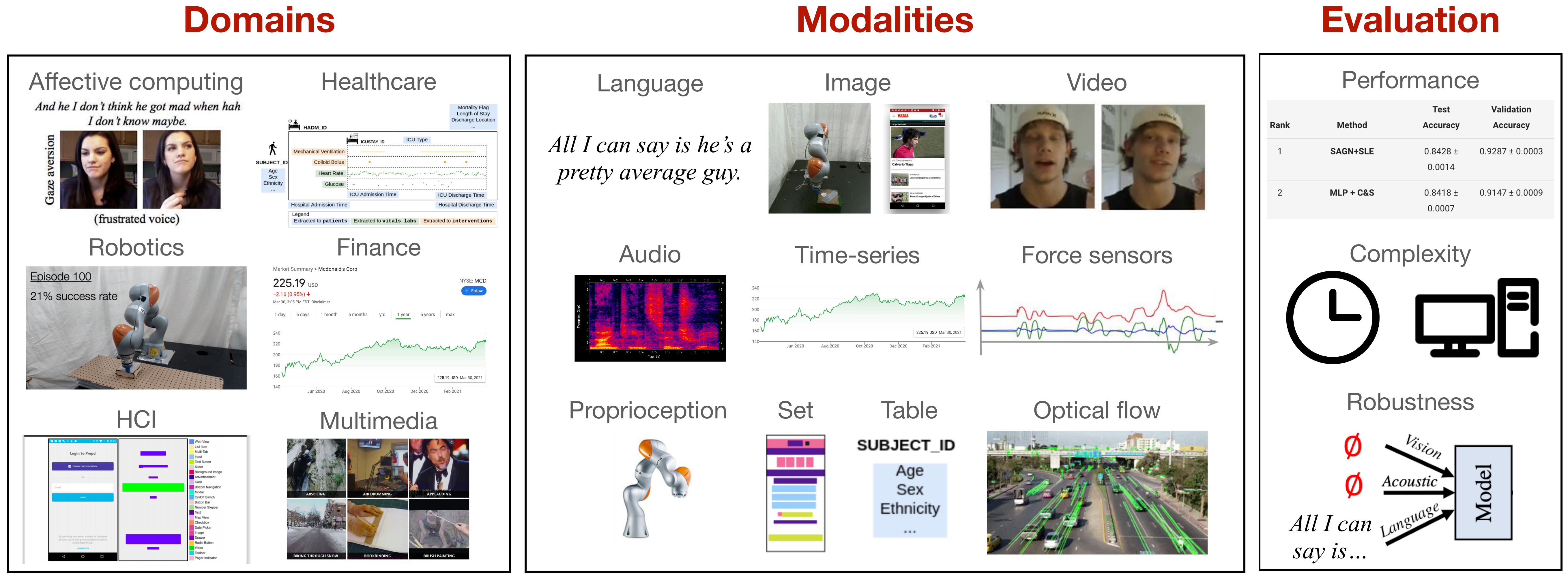}
\vspace{-0mm}
\caption{\names\ contains a diverse set of $15$ datasets spanning $10$ modalities and testing for more than $20$ prediction tasks across $6$ distinct research areas, and enables standardized, reliable, and reproducible large-scale benchmarking of multimodal models for performance, complexity, and robustness.\vspace{-4mm}}
\label{figs:overview}
\end{figure*}

The research field of multimodal machine learning (ML) brings unique challenges for both computational and theoretical research given the heterogeneity of various data sources~\citep{baltruvsaitis2018multimodal,liang2022foundations}. At its core lies the learning of \textit{multimodal representations} that capture correspondences between modalities for prediction, and has emerged as a vibrant interdisciplinary field of immense importance and with extraordinary potential in multimedia~\citep{1667983,liang2023highmodality}, affective computing~\citep{liang2019tensor,PORIA201798}, robotics~\citep{kirchner2019embedded,lee2019making}, finance~\citep{doi:10.1177/0170840618765019}, dialogue~\citep{Pittermann2010}, human-computer interaction~\citep{dumas2009multimodal,obrenovic2004modeling}, and healthcare~\citep{medical,xu2019multimodal}. In order to accelerate research in building general-purpose multimodal models across diverse research areas, modalities, and tasks, we contribute \names\ (Figure~\ref{figs:overview}), a systematic and unified large-scale benchmark that brings us closer to the requirements of real-world multimodal applications. \names\ contains a diverse set of $15$ datasets spanning $10$ modalities and testing for $20$ prediction tasks across $6$ distinct research areas, and is designed to comprehensively evaluate generalization across domains and modalities, complexity during training and inference, and robustness to noisy and missing modalities. Additionally, we release \codes, a public toolkit consisting of standardized implementations of $>20$ core multimodal algorithms in a modular fashion to enable accessibility for new researchers, compositionality of approaches, and reproducibility of results. Together, these public resources ensure ease of use, accessibility, and reproducibility, and they will be continually expanded in courses, workshops, and competitions around the world.

\begin{figure*}[]
\centering
\vspace{0mm}
\includegraphics[width=0.9\linewidth]{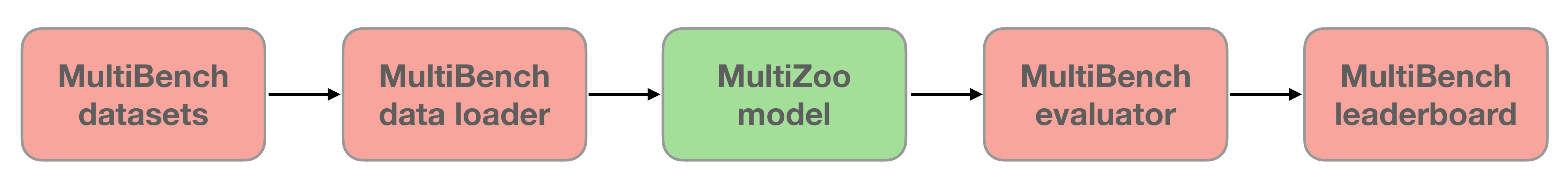}
\vspace{-0mm}
\caption{Our \names\ toolkit provides a machine learning pipeline across data processing, data loading, multimodal models, evaluation metrics, and a public leaderboard to encourage accessible, standardized, and reproducible research in multimodal representation learning.\vspace{-4mm}}
\label{figs:pipeline}
\end{figure*}

\vspace{-3mm}
\section{\names\ and \codes}
\vspace{-2mm}

\names\ provides a standardized machine learning pipeline that starts from data loading to running multimodal models, providing evaluation metrics, and a public leaderboard to encourage future research in multimodal representation learning (see Figure~\ref{figs:pipeline}).

\textbf{\names\ datasets:} Table~\ref{data:overview} shows an overview of the datasets provided in \names, which span research areas in multimedia, affective computing, robotics, finance, human-computer interaction, and healthcare, more than $15$ datasets, $10$ modalities, and $20$ prediction tasks.

\begin{table*}[]
\fontsize{9}{11}\selectfont
\setlength\tabcolsep{3.0pt}
\vspace{-4mm}
\caption{\names\ provides a comprehensive suite of $15$ datasets covering a diverse range of $6$ research areas, dataset sizes, $10$ input modalities (in the form of $\ell$: language, $i$: image, $v$: video, $a$: audio, $t$: time-series, $ta$: tabular, $f$: force sensor, $p$: proprioception sensor, $s$: set, $o$: optical flow), and $20$ prediction tasks.}
\centering
\footnotesize
\vspace{-2mm}
\begin{tabular}{l|lccccccc}
\Xhline{3\arrayrulewidth}
\multicolumn{1}{l|}{Research Area} & Size & Dataset & Modalities & \# Samples & Prediction task &   \\
\Xhline{0.5\arrayrulewidth}
\multirow{4}{*}{Affective Computing} & \multirow{1}{*}{S} & \textsc{MUStARD}~\citep{castro2019towards} & $\{\ell,v,a\}$ & $690$ & sarcasm \\
& \multirow{1}{*}{M} & \textsc{CMU-MOSI}~\citep{zadeh2016mosi} & $\{\ell,v,a\}$ & $2,199$ & sentiment \\
& \multirow{1}{*}{L} & \textsc{UR-FUNNY}~\citep{hasan2019ur} & $\{\ell,v,a\}$ & $16,514$ & humor \\
& \multirow{1}{*}{L} & \textsc{CMU-MOSEI}~\citep{zadeh2018multimodal} & $\{\ell,v,a\}$ & $22,777$ & sentiment, emotions \\
\Xhline{0.5\arrayrulewidth}
\multirow{1}{*}{Healthcare} & L & \textsc{MIMIC}~\citep{MIMIC} & $\{t,ta\}$ & $36,212$ & mortality, ICD-$9$ codes\\
\Xhline{0.5\arrayrulewidth}
\multirow{2}{*}{Robotics} & M & \textsc{MuJoCo Push}~\citep{lee2020multimodal} & $\{i,f,p\}$ & $37,990$ & object pose \\ 
& L & \textsc{Vision\&Touch}~\citep{lee2019making} & $\{i,f,p\}$ & $147,000$ & contact, robot pose \\
\Xhline{0.5\arrayrulewidth}
\multirow{3}{*}{Finance} & M & \textsc{Stocks-F\&B} & $\{t \times 18\}$ & $5,218$ & stock price, volatility \\
& M & \textsc{Stocks-Health} & $\{t \times 63\}$ & $5,218$ & stock price, volatility \\
& M & \textsc{Stocks-Tech} & $\{t \times 100\}$ & $5,218$ & stock price, volatility \\
\Xhline{0.5\arrayrulewidth}
\multirow{1}{*}{HCI} & S & \textsc{ENRICO}~\citep{leiva2020enrico} & $\{i,s\}$ & $1,460$ & design interface \\
\Xhline{0.5\arrayrulewidth}
\multirow{3}{*}{Multimedia} & \multirow{1}{*}{M} & \textsc{MM-IMDb}~\citep{arevalo2017gated} & $\{\ell,i\}$ & $25,959$ & movie genre \\
& \multirow{1}{*}{M} & \textsc{AV-MNIST}~\citep{vielzeuf2018centralnet} & $\{i,a\}$ & $70,000$ & digit \\
& \multirow{1}{*}{L} & \textsc{Kinetics400}~\citep{kay2017kinetics} & $\{v,a,o\}$ & $306,245$ & human action \\
\Xhline{3\arrayrulewidth}
\end{tabular}
\vspace{-2mm}
\label{data:overview}
\end{table*}

\begin{figure*}[tbp]
\centering
\vspace{-2mm}
\includegraphics[width=0.95\linewidth]{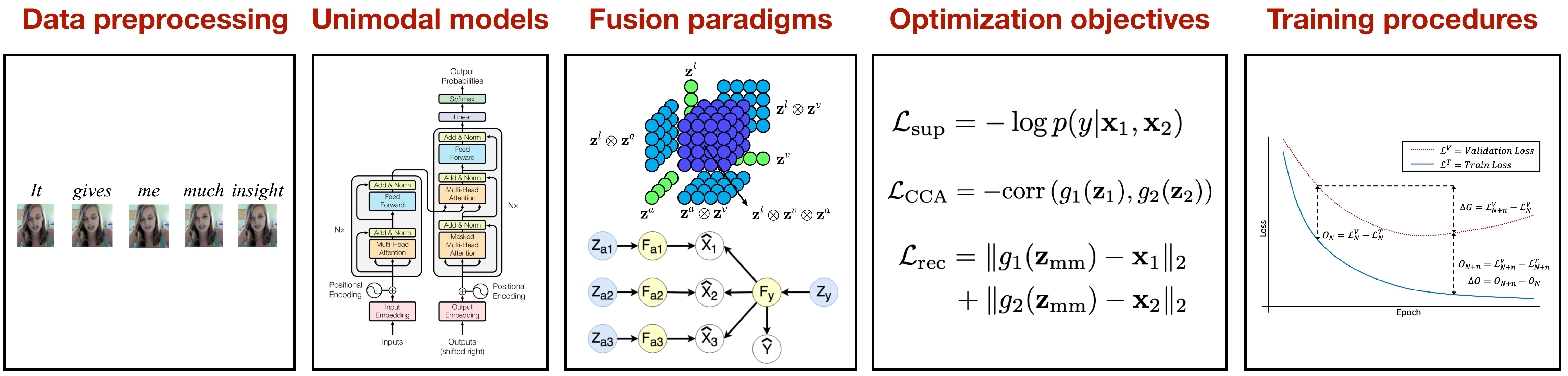}
\vspace{-2mm}
\caption{\codes\ provides a standardized implementation of multimodal methods in a modular fashion to enable accessibility for new researchers, compositionality of approaches, and reproducibility of results.}
\label{figs:multizoo}
\vspace{-1mm}
\end{figure*}

\textbf{\codes: A zoo of multimodal algorithms:} To complement \names, we release a comprehensive toolkit, \codes, as starter code for multimodal algorithms which implements $20$ methods spanning different methodological innovations in (1) data preprocessing, (2) fusion paradigms, (3) optimization objectives, and (4) training procedures (see Figure~\ref{figs:multizoo}). Each of these algorithms are chosen because they provide unique perspectives to the technical challenges in multimodal learning~\citep{baltruvsaitis2018multimodal} (see Table~\ref{data:models} for details).

\textbf{Evaluation protocol:} \names\ contains evaluation scripts for the following holistic desiderata in multimodal learning:
(1) Performance: We standardize evaluation using MSE and MAE for regression, as well as accuracy, micro \& macro F1-score, and AUPRC for classification.
(2) Complexity: We record the amount of information taken in bits (i.e., data size), the number of model parameters, as well as time and memory resources required during the entire training process. Real-world models may also need to be small and compact to run on mobile devices~\citep{radu2016towards} so we also report inference time and memory on CPU and GPU. The datasets and models included are designed to span a range of compute times from 1 minute to 6 hours, memory from 2GB to 12GB, models from 0.01 million to 280 million parameters, and datasets from 690 to 147,000 samples.
(3) Robustness: The toolkit includes both \textit{modality-specific imperfections} taking into account each modality's unique noise topologies (i.e., flips and crops of images, natural misspellings in text, abbreviations in spoken audio), and \textit{multimodal imperfections} across modalities (e.g., missing modalities, or a chunk of time missing in time-series data)~\citep{liang2019tensor,pham2019found}.

\textbf{Installation, testing, and integration:} Our documentation provides installation instructions in Linux, MacOS, and Windows. We also include a suite of unit tests (testing self-contained functions) and integration tests (testing multiple components from across the unimodal, fusion, and training loop modules together) with $100\%$ coverage for self-contained functions and $88\%$ coverage overall including integration tests. We also include instructions for continuous integration: our software is hosted on GitHub which enables version control and integration via pull requests and merges. We enabled GitHub Actions workflows, which automatically runs the test builds and is triggered every time new changes are incorporated. After making the desired changes and making sure all tests pass, users can create a pull request and the authors will merge these changes into the main branch.

\textbf{Together:} In Algorithm~\ref{alg:code}, we show a sample code snippet in Python that loads a dataset, defines the unimodal and multimodal architectures, optimization objective, and training procedures, before running the evaluation protocol. Our toolkit is easy to use and trains models in less than $10$ lines of code. By standardizing the implementation of each module and disentangling individual modules, optimizations, and training, \codes\ ensures accessibility and reproducibility of its algorithms.

\begin{table*}[]
\fontsize{9}{11}\selectfont
\setlength\tabcolsep{1.0pt}
\vspace{-2mm}
\caption{\codes\ provides a standardized implementation of the following multimodal methods spanning data processing, fusion paradigms, optimization objectives, and training procedures, which offer complementary perspectives towards tackling multimodal challenges in alignment, complementarity, and robustness.}
\centering
\footnotesize
\vspace{-0mm}

\begin{tabular}{c|c|c|c|c}
\Xhline{3\arrayrulewidth}
Category & Method & Alignment & Complementarity & Robustness \\
\Xhline{0.5\arrayrulewidth}
\multirow{1}{*}{Data} & \textsc{WordAlign}~\citep{chen2017multimodal} & \textcolor{gg}\cmark & \textcolor{rr}\xmark & \textcolor{rr}\xmark \\
\Xhline{0.5\arrayrulewidth}
\multirow{6}{*}{Model} & \textsc{EF}, \textsc{LF}~\citep{baltruvsaitis2018multimodal} & \textcolor{rr}\xmark & \textcolor{gg}\cmark & \textcolor{rr}\xmark \\
& \textsc{TF}~\citep{zadeh2017tensor}, \textsc{LRTF}~\citep{liu2018efficient} & \textcolor{rr}\xmark & \textcolor{gg}\cmark & \textcolor{rr}\xmark \\
& \textsc{MI-Matrix}, \textsc{MI-Vector}, \textsc{MI-Scalar}~\citep{Jayakumar2020Multiplicative} & \textcolor{rr}\xmark & \textcolor{gg}\cmark & \textcolor{rr}\xmark \\
& \textsc{NL Gate}~\citep{wang2020makes} & \textcolor{rr}\xmark & \textcolor{gg}\cmark & \textcolor{rr}\xmark \\
& \textsc{MulT}~\citep{tsai2019multimodal} & \textcolor{gg}\cmark & \textcolor{gg}\cmark & \textcolor{rr}\xmark \\
& \textsc{MFAS}~\citep{perez2019mfas} & \textcolor{rr}\xmark & \textcolor{gg}\cmark & \textcolor{rr}\xmark \\
\Xhline{0.5\arrayrulewidth}
\multirow{5}{*}{Objective} & \textsc{CCA}~\citep{andrew2013deep} & \textcolor{gg}\cmark & \textcolor{rr}\xmark & \textcolor{rr}\xmark \\
& \textsc{ReFNet}~\citep{sankaran2021multimodal} & \textcolor{gg}\cmark & \textcolor{rr}\xmark & \textcolor{rr}\xmark \\
& \textsc{MFM}~\citep{tsai2019learning} & \textcolor{rr}\xmark & \textcolor{gg}\cmark & \textcolor{rr}\xmark \\
& \textsc{MVAE}~\citep{wu2018multimodal} & \textcolor{rr}\xmark & \textcolor{gg}\cmark & \textcolor{rr}\xmark \\
& \textsc{MCTN}~\citep{pham2019found} & \textcolor{rr}\xmark & \textcolor{rr}\xmark & \textcolor{gg}\cmark \\
\Xhline{0.5\arrayrulewidth}
\multirow{2}{*}{Training} & \textsc{GradBlend}~\citep{wang2020makes} & \textcolor{rr}\xmark & \textcolor{gg}\cmark & \textcolor{gg}\cmark \\
& \textsc{RMFE}~\citep{gat2020removing} & \textcolor{rr}\xmark & \textcolor{gg}\cmark & \textcolor{gg}\cmark \\
\Xhline{0.5\arrayrulewidth}
\Xhline{3\arrayrulewidth}
\end{tabular}

\vspace{-2mm}
\label{data:models}
\end{table*}


\newlength{\textfloatsepsave} \setlength{\textfloatsepsave}{\textfloatsep} \setlength{\textfloatsep}{4mm}

\begin{algorithm}[tb]
    \caption{PyTorch code integrating \names\ datasets and \codes\ models.}
    \label{alg:code}
   
    \definecolor{codeblue}{rgb}{0.25,0.5,0.5}
    \lstset{
      basicstyle=\fontsize{7.2pt}{7.2pt}\ttfamily\bfseries,
      commentstyle=\fontsize{7.2pt}{7.2pt}\color{codeblue},
      keywordstyle=\fontsize{7.2pt}{7.2pt},
    }
\begin{lstlisting}[language=python]
from datasets.get_data import get_dataloader
from unimodals.common_models import ResNet, Transformer
from fusions.common_fusions import MultInteractions
from training_structures.gradient_blend import train, test

# load Multimodal IMDB dataset
traindata, validdata, testdata = get_dataloader('multimodal_imdb')
out_channels = 3
# define ResNet and Transformer unimodal encoders
encoders = [ResNet(in_channels=1, out_channels=3, layers=5),
            Transformer(in_channels=1, out_channels=3, layers=3)]
# define a Multiplicative Interactions fusion layer
fusion = MultInteractions([out_channels*8, out_channels*32], out_channels*32, 'matrix')
classifier = MLP(out_channels*32, 100, labels=23)
# train using Gradient Blend algorithm
model = train(encoders, fusion, classifier, traindata, validdata, 
        epochs=100, optimtype=torch.optim.SGD, lr=0.01, weight_decay=0.0001)
# test
performance, complexity, robustness = test(model, testdata)
\end{lstlisting}
\vspace{-2mm}
\end{algorithm}

\vspace{-3mm}
\section{Results}
\vspace{-2mm}

\codes\ and \names\ enable quick experimentation of multimodal algorithms for performance while balancing complexity and robustness. They uncover several shortcomings of current models, including poor generalization to out-of-domain tasks, tradeoffs between performance and efficiency, and lack of robustness to real-world imperfections. Our resources also pave the way toward answering novel research questions in multimodal transfer learning, multi-task learning, co-learning, pre-training, and interpretability. We include these results and discussions in our full paper~\citep{liang2021multibench} as well as scripts to reproduce these results in \names\ software.

\vspace{-3mm}
\section{Conclusion}
\vspace{-1mm}

In conclusion, we present \codes\ and \names, a large-scale open-source toolkit unifying previously disjoint efforts in multimodal research with a focus on ease of use, accessibility, and reproducibility, thereby enabling a deeper understanding of multimodal models. Through its unprecedented range of research areas, datasets, modalities, tasks, and evaluation metrics, our toolkit paves the way toward building more generalizable, lightweight, and robust multimodal models. These tools have already been used for new directions for visualizing trained models~\citep{liang2023multiviz}, large-scale multimodal foundation models~\citep{liang2023highmodality}, multimodal fusion methods~\citep{huang2022modality,xue2023dynamic}, and other theoretical and empirical studies of multimodal learning in applications ranging from robotics~\citep{li2022see} and HCI~\citep{wu2023webui} to IoT~\citep{hou2023architecting}, remote sensing~\citep{xiong2022interpretable}, and healthcare~\citep{suvon2022multimodal}. Our toolkits are publicly available, regularly updated with new tasks and modeling paradigms, and welcome inputs from the community.

{\footnotesize
\bibliography{refs}
\bibliographystyle{plain}
}

\end{document}